\documentclass[conference]{IEEEtran}
\IEEEoverridecommandlockouts
\usepackage{cite}
\usepackage{amsmath,amssymb,amsfonts}
\usepackage{algorithmic}
\usepackage{graphicx}
\usepackage{textcomp}
\usepackage{xcolor}
\usepackage{graphicx}
\usepackage{subcaption}
\def\BibTeX{{\rm B\kern-.05em{\sc i\kern-.025em b}\kern-.08em
    T\kern-.1667em\lower.7ex\hbox{E}\kern-.125emX}}
\begin{document}

\title{Electricity Demand Forecasting through Natural Language Processing with Long Short-Term Memory Networks\\
\thanks{The author Yun BAI was supported by the program of the China Scholarship Council (CSC Nos. 202106020064).}
}

\author{\IEEEauthorblockN{Yun Bai}
\IEEEauthorblockA{\textit{Centre PERSEE} \\
\textit{Mines Paris - PSL}\\
Sophia Antipolis, France \\
yun.bai@minesparis.psl.eu}
\and
\IEEEauthorblockN{Simon Camal}
\IEEEauthorblockA{\textit{Centre PERSEE} \\
\textit{Mines Paris - PSL}\\
Sophia Antipolis, France \\
simon.camal@minesparis.psl.eu}
\and
\IEEEauthorblockN{Andrea Michiorri}
\IEEEauthorblockA{\textit{Centre PERSEE} \\
\textit{Mines Paris - PSL}\\
Sophia Antipolis, France \\
andrea.michiorri@minesparis.psl.eu}
}

\maketitle

\begin{abstract}
Electricity demand forecasting is a well established research field. Usually this task is performed considering historical loads, weather forecasts, calendar information and known major events. Recently attention has been given on the possible use of new sources of information from textual news in order to improve the performance of these predictions. 
This paper proposes a Long and Short-Term Memory (LSTM) network incorporating textual news features that successfully predicts the deterministic and probabilistic tasks of the UK national electricity demand.
The study finds that public sentiment and word vector representations related to transport and geopolitics have time-continuity effects on electricity demand. 
The experimental results show that the LSTM with textual features improves by more than 3\% compared to the pure LSTM benchmark and by close to 10\% over the official benchmark.
Furthermore, the proposed model effectively reduces forecasting uncertainty by narrowing the confidence interval and bringing the forecast distribution closer to the truth.

\end{abstract}


\section{Introduction}
\label{Introduction}
Electricity demand forecasting, as a decision process in the electricity market, is an essential step in network operation \cite{fallah2018computational}. 
Precise forecasting of electricity demand not only assists the operators in allocating efficiently resources but benefits the safety assessment in energy systems \cite{metaxiotis2003artificial}. 
The past few decades have witnessed the development of research on electricity demand forecasting. 
Early techniques, such as linear and non-parameter regression, perform well to some extent while facing difficulty dealing with changes in meteorology, society, economy, and policy.
Especially with the combination of smart grid and electricity systems, and the penetration of renewable resources, demand forecasting gets more complex than before. 
Artificial intelligent methods regard the elements that influence electricity demand and capture the non-linear relationships between historical data and external variables. 
Researchers aim to build intelligent, adaptive, and optimal energy systems through AI. 
Examples of widely used AI methods for demand forecasting are support vector machines, artificial neural networks, self-organized maps, extreme learning machines, and so on \cite{fallah2018computational}.

Besides the efforts put forward on developing new models or re-construction of model structures, researchers are still investigating information that benefits the forecasting model. 
One typical way is to select several historical lags via Akaike Information Criterion (AIC) or the Bayesian Information Criterion (BIC) as in \cite{yang2019short}. 
This feature selection method may sometimes suit machine learning models but needs more time sequence structure. 
Another source of information comes from statistical data sharing. 
In hierarchical forecasting, \cite{nystrup2020temporal} designed four estimators for forecasting the electricity demand in different regions in Sweden. 
The auto-covariance matrices, auto-correlation and variance, paired cross-correlation matrices, and reliable estimation of inverse matrices within the aggregated demand levels were considered to improve information sharing. 
Besides, one should pay attention to the ability of meteorology and calendar data in demand forecasting.
Evidence shows that electricity demand is a time series with solid seasonal signals. 
The benchmark models often include the meteorology and calendar features in the research nowadays \cite{thorey2018ensemble,browell2021probabilistic,sgarlato2022role}.


There is a growing trend of information fusion to improve traditional models, such as text-based forecasting. 
With the development of the internet, people can quickly post their comments and views online. 
The internet is full of noisy, unstructured, and sparse knowledge, the distribution of which is of difficult organisation \cite{smith2010text}. 
However, Natural Language Processing (NLP) offers the chance to analyze such information and connect text and events through statistical models. 
In state-of-the-art research, news reports, online search traffics, social media, knowledge forums, books, and policies are usually served as the corpus sources \cite{li2019text,wu2021effective,zhou2016can,schaer2022predictive,wang2023forecasting,kozlowski2019geometry}. 
Various forms of text features can be used as external variables for forecasting models, such as the quantitative sentiments by TextBlob, topic distributions through Latent Dirichlet Allocation (LDA) \cite{bai2022crude}, the deep-processed variables by Convolutional Neural Network (CNN) \cite{wu2021effective}, the word embeddings grammthrough by pre-trained large language models, the knowledge graph by extraction of items and relationships \cite{sawhney2020multimodal,rodrigues2019combining,liu2019combining}. 
Text-based forecasting now is matured in the fields of crude oil price \cite{bai2022crude}, financial risk \cite{sawhney2020multimodal},  health insurance \cite{kreif2021estimating}, and movie revenues \cite{joshi2010movie}.

In electricity demand forecasting, text-based forecasting has emerged as a possible alternative in the last years.
In \cite{obst2019textual}, the authors considered the use of weather reports texts as the supplementary in the absence of weather data.
They quantified the effect of word frequency on forecasting and found that the word embedding vectors had geographical properties.
This research is a beneficial attempt at forecasting with text information, although the external information only marginally improved.
In the following research, the authors continued to add a data source from Twitter that was expected to offer more elements about society and economy \cite{obst2021textual}.
They treated the number of tweets containing `télétravail' (French for 'teleworking') and its variants as features to correct the residuals generated by Generalized Additive Models (GAM) model.
The results showed statistical improvement in the benchmark model and depicted the demand change after the lockdown.
The research in \cite{wang2023forecasting} illustrates another example of incorporating text information to demand forecast.
The authors explored improving the forecasting of Chinese monthly electricity consumption with word embedding vectors extracted by the CNN module.

The above studies demonstrate the potential of text information as an essential supplement.
Although they made progress in electricity demand forecasting, there is still room for improvement.
A previous work by the authors has tried to both explore alternative methods to word frequencies and to trey to explain the mechanisms linking news and electricity demand highlighted by the improved model \cite{bai2023quantitative}. In this work the authors explored how the five types of text features (count, word frequency, sentiment, topic distribution, word embedding) from news titles, descriptions, and text bodies influence forecasting. It was also explained the improvement brought by text from views of global and local correlations, and causality effects. The conclusion was that keywords related to major social events, the minimum subjectivity of sentiments, and the word embedding dimensions of international conflicts benefit the forecasting. This effect is not due to coincidence.

This paper extends and broadens the scenario for the study of \cite{bai2023quantitative}.
The main objective of the work is to verify the generalization and transferability of text information under another forecasting paradigm.
Concretely, we considered the sustained influence of news text instead of forecasting with the news from the previous day.

The contributions of this paper are as follows:
\begin{itemize}
    \item To verify the influence of persistent trends in textual-based features
    \item To explore the performance of such method in probabilistic forecasting 
\end{itemize}

This is done thanks to the following steps:
\begin{itemize}
    \item We built Long Short-Term Memory (LSTM) networks to keep the news memories for at least one week and identified the text features with sustained influence.
    \item We developed a dimension reduction network through CNN autoencoder for hundreds of features.
    \item We proved by experiments that text information enhanced both deterministic and probabilistic forecasting.
\end{itemize}

In this paper, Section~\ref{Introduction} introduces the research background and a short literature review. Section~\ref{Methods} illustrates the methods used, including the forecasting framework, the datasets used and the evaluation metrics. The experimental setup and results are shown in Section~\ref{Experiments}, with an analysis of the improvement. Section~\ref{Conclusions} concludes the paper.

\section{Methods}
\label{Methods}
\subsection{Research framework}
\begin{figure*}[ht]
\centering
\centerline{\includegraphics[scale=.58]{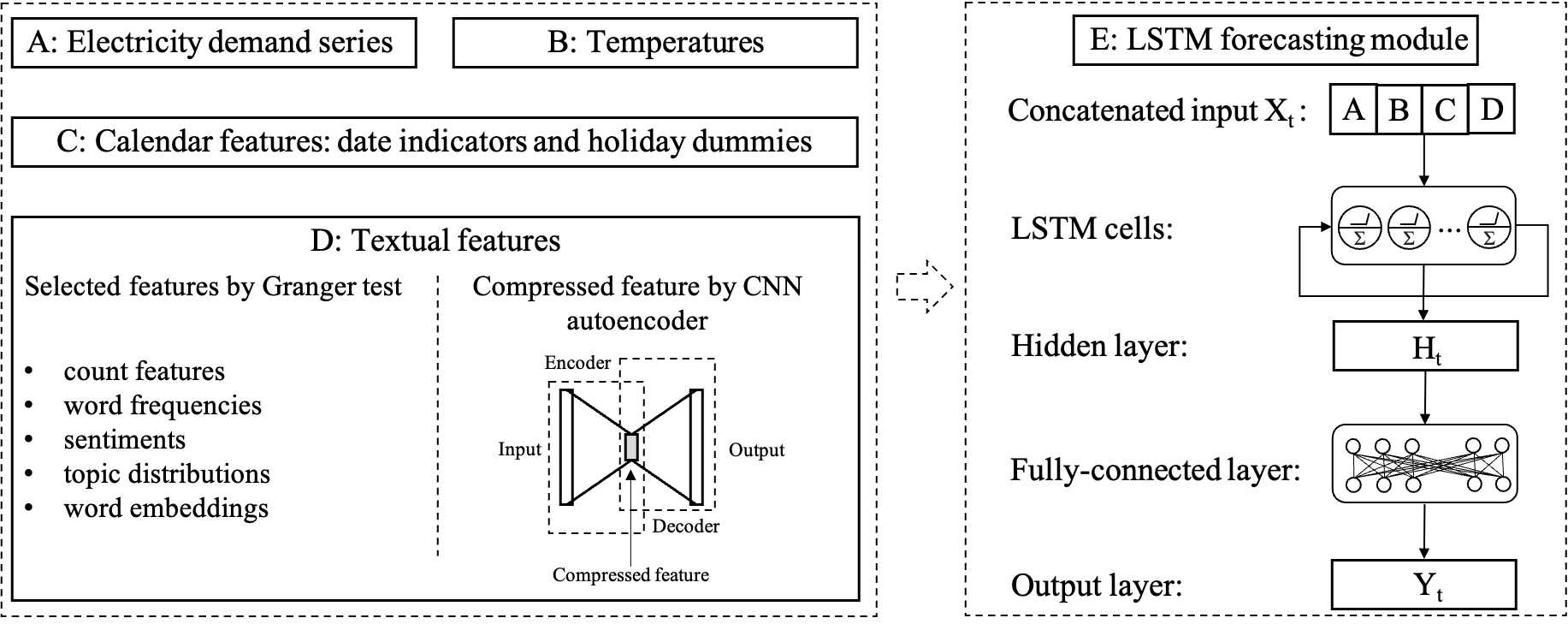}}
\caption{The forecasting framework of this research.}
\label{workflow}
\end{figure*}

The research framework has been summarised in Figure~\ref{workflow}, mainly including the data acquisition and feature pre-processing (left) and forecasting model (right).

This paper preserves the research case from \cite{bai2023quantitative} with the datasets.
Electricity demand is collected in the UK region from ENTSO-E transparency platform \cite{entso-e} as in block A.
We take the temperature observations (block B) and historical bank holidays (stored as dummy variables in block C) from London.
The reason of using the bank holiday is to model the relationships of human activities and load demand.
Block C contains also date indicators, important for the seasonality in electricity demand. These are encoded by the sine and cosine of one day in a week and year.
Textual data used come from the British Broadcasting Corporation (BBC) \cite{BBC_news}, which serves the corpus in block D.
All the datasets are from the same period (01/06/2016 - 31/05/2021), and we treat the first four years as a training set and the data of last year as a test set.

In block D, the textual features contain five groups: count features, word frequencies, sentiments, topic distributions, and word embeddings, whose definitions and achievements could be referred to in \cite{bai2023quantitative}.
This paper uses two treatments of the features.
Firstly, a Granger test is used in order to reduce the number of used features. 
Secondly, a CNN autoencoder is built as another dimension compression machine to extract representations from high-dimensional textual features \cite{michelucci2022introduction}.
In detail, the CNN autoencoder compresses and decompresses data through the encoder and decoder architectures.
We minimize the loss function between the input and output, and reserve the representation in the hidden layer as the compressed feature.
Respect to a traditional dimension reduction methods, such as Principal Component Analysis (PCA), CNN autoencoder can capture non-linear and more complex patterns from inputs.
Besides, as an unsupervised learning method, it can extract a global representation without data labeling.

The demand data and the corresponding features are concatenated as tensors to input the forecasting model in block E.
In the forecasting module, the inputs are first fed into the LSTM architecture with several cells \cite{hochreiter1997long}.
The input gate of LSTM can learn to recognize the critical information and store it in the long-term state.
The forget gate learns to keep and extract hidden information through on-demand retrieval.
The output gate controls the information from the long-term state and outputs the short-term state at the current time step.
Finally, in our module, the output in LSTM is mapped through a fully-connected layer to the shape of the forecasting horizon as the final predictions.
In our case, we set the number of neurons less than the forecasting horizon to reduce the computational complexity and better extract the long-term dependencies in the inputs.

\subsection{Evaluation}
The performance of the forecasting model with textual features is evaluated through deterministic and probabilistic metrics.
In deterministic evaluation, Root Mean Squared Error (RMSE), Mean Absolute Error (MAE), and Symmetric Mean Absolute Percentage Error (SMAPE) are used as metrics. Formulae for these common metrics are here omitted for space reasons.

For the evaluation of probabilistic forecasts the following metrics are considered: Pinball loss, as in (\ref{pinball loss}), Winkler score, as in (\ref{Winkler}), and Continuous Ranked Probability Score (CRPS) as in (\ref{CRPS equation}) \cite{Rob2021Evaluating}.

\begin{equation}
\label{pinball loss}
\operatorname{P_\rho} = \frac{1}{N}\sum_{i=1}^N\operatorname{max}((\rho - 1)\cdot(y_i-\hat{y_i})).
\end{equation}

\begin{equation}
\label{Winkler}
\operatorname{W_{\alpha,i}} = 
\begin{cases}
(u_{\alpha,i}-l_{\alpha,i}) + \frac{2}{\alpha}(l_{\alpha,i}-\hat{y_i}), &  \hat{y_i} < l_{\alpha,i} \\
\newline
(u_{\alpha,i}-l_{\alpha,i}), &  l_{\alpha,i} \leq \hat{y_i} \leq u_{\alpha,i} \\
\newline
(u_{\alpha,i}-l_{\alpha,i}) + \frac{2}{\alpha}(\hat{y_i} - u_{\alpha,i}), & \hat{y_i} > u_{\alpha,i},
\end{cases}
\end{equation}

where \emph{N} is the number of samples, $y_i$ and $\hat{y_i}$ are truth and prediction, $[l_{\alpha,i},u_{\alpha,i}]$ is the $100(1-\alpha)\%$ prediction interval, and $W_{\alpha,i}$ is the length of the interval and a penalty value if $\hat{y_i}$ falls out of $[l_{\alpha,i},u_{\alpha,i}]$.
We use the average of the $W_{\alpha,i}$ to measure the performance on a sample, that is $W_\alpha = \frac{1}{N}\sum_{i=1}^{N}W_{\alpha,i}$.

\begin{equation}
\label{CRPS equation}
    \operatorname{CRPS_{i}}(\hat{y_i},y_i) = \int_{-\infty}^{\infty} (\operatorname{CDF}(x|\hat{y_i},\sigma) -\operatorname{H}(x-y_i))^2 dx.
\end{equation}

(\ref{CRPS equation}) computes the CRPS value for a single forecast-truth pair by comparing the Probability Distribution Function (PDF) of the forecast to the truth.
$\operatorname{CDF}(x|\hat{y_i},\sigma)$ is the Cumulative Distribution Function (CDF) of the forecasts, where we suppose the CDF is a normal distribution with the mean of $\hat{y_i}$ and standard deviation of $\sigma$.
$\operatorname{H}(x-y_i)$ represents the Heaviside step function.
$\operatorname{H}(x-y_i) = 0$ when $x<y_i$, and $\operatorname{H}(x-y_i) = 1$ otherwise.
We use the $\operatorname{CRPS} = \frac{1}{N}\sum_{i=1}^{N} \operatorname{CRPS_i}$ to evaluate the performance on a sample.

\section{Results}
\label{Experiments}
\subsection{Model configurations}
We still focus on the day-ahead forecasting of electricity demand data. On day $D$, we use the demand lags from the past week with $D$ excluded to forecast the demand leads on the day $D+1$.
The resolution of the demand data is half an hour; thus, the length of lags equals 336, and the forecasting horizons are from 48 to 96 time steps ahead.

For the CNN autoencoder, we set the input and output channels equal to the number of textual features in a specific group, and only keep the hidden compressed state into 1.
For example, we build a three-layer autoencoder for the 100 features in the group `Word embedding'.
The encoder first compresses the 100-dimension word vectors into 1d, then the decoder recovers the hidden 1d into 100d again.
The LSTM architecture contains one layer with 24 neurons, and the fully-connected layer maps the 24d output into 48d.
The loss function of CNN autoencoder and LSTM is Mean Squared Error (MSE), and we change the loss function of LSTM into Pinball loss when we turn to quantile forecasting.
The batch size is set to 4, the learning rate is 1e-4, the optimizer is Adam, and an early stopping mechanism controls the training process.

\subsection{Deterministic forecasting results}
In this section, we first build a benchmark LSTM model, including the external features of temperatures, date indicators, and holidays.
Afterward, we conduct add the five groups of textual features, as shown in Figure~\ref{workflow}.
Moreover, we continue to improve the model by integrating the compressed features and finally evaluate the text-based model from the deterministic view with daily averaged metrics.
The results are illustrated in Table~\ref{Deterministic-results-comparision}.

\begin{table}[ht]
\centering
\caption{Deterministic results comparison}
\label{Deterministic-results-comparision}
\scalebox{1.2}{
\begin{tabular}{lrrr}
\hline
Models     & RMSE    & MAE     & SMAPE(\%) \\ \hline
ENTSO-E        & 2800.50 & 2544.86 & 7.65      \\
ExtraTree & 2800.77 & 2374.07 & 7.29      \\
ExtraTree-Text & \textbf{2684.62} & 2263.86 & 6.92      \\
\hline
LSTM & 2775.99 & 2333.20 & 7.10 \\
LSTM-W-S-T-G & 4853.01 & 4094.80 & 12.39 \\
LSTM-S-G & 2732.44 & 2299.20 & 6.99\\
LSTM-S-G-CG     & 2692.33 & \textbf{2248.55} & \textbf{6.83}      \\ \hline
\end{tabular}}
\end{table}

In Table~\ref{Deterministic-results-comparision}, the `ENTSO-E' is the official day-ahead forecasting benchmark.
The `ExtraTree' is the benchmark model from our previous work.
For forecasting with textual features, the `ExtraTree-Text' is the best-performing model from \cite{bai2023quantitative}, which includes the textual features of words frequencies from news titles, the sentiments and GloVe word embeddings from news text bodies.

The `ENTSO-E', `ExtraTree', and `ExtraTree-Text' are cited here as the baselines of this research.
In this study, we carry out the ablation experiments by adding the textual features into the LSTM benchmark.
We only include three set of results in Table~\ref{Deterministic-results-comparision} for the limitation of pages.
We note the word frequencies, sentiments, topics, and GloVe word embeddings as \textbf{W}, \textbf{S}, \textbf{T}, \textbf{G}, and the compressed embeddings by CNN autoencoder as \textbf{CG}.
The results reveal that the inclusion of word frequencies and topics did not yield significant improvements, whereas the compressed word embeddings led to a notable enhancement.

\begin{figure}[ht]
    \centering
    \centerline{\includegraphics[scale=0.1]{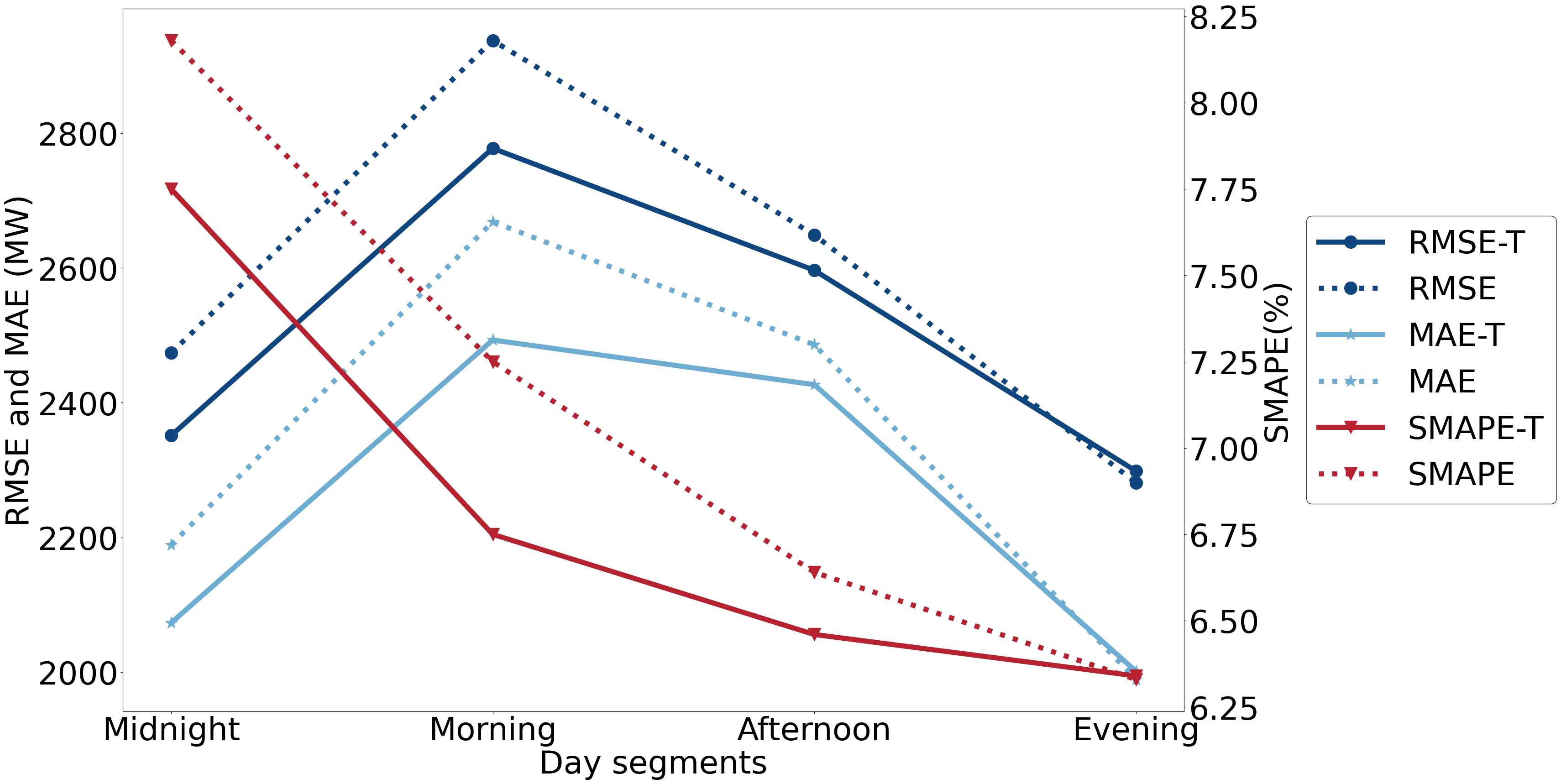}}
    \caption{The evaluations of the LSTM model with and without textual features in different day segments. The dashed lines represent the forecasting with only LSTM model, and the solid lines are the forecasting with textual features.}
    \label{ImprovementDay}
\end{figure}

Additionally, we group the forecasting errors of different hours within a day and split a day into several segments: midnight (1h-6h), morning (7h-12h), afternoon (13h-18h), and evening (19h-24h).
The comparisons between the LSTM with and without textual features are shown in Figure~\ref{ImprovementDay}.
The textual features enhance the LSTM benchmark in the morning especially, with the improvement of more than 5\%, 6\%, and 7\% on RMSE, MAE, and SMAPE.

\subsection{Probabilistic forecasting results}
\begin{figure*}[ht]
\centering
\begin{subfigure}{0.3\textwidth}
  \includegraphics[width=\linewidth]{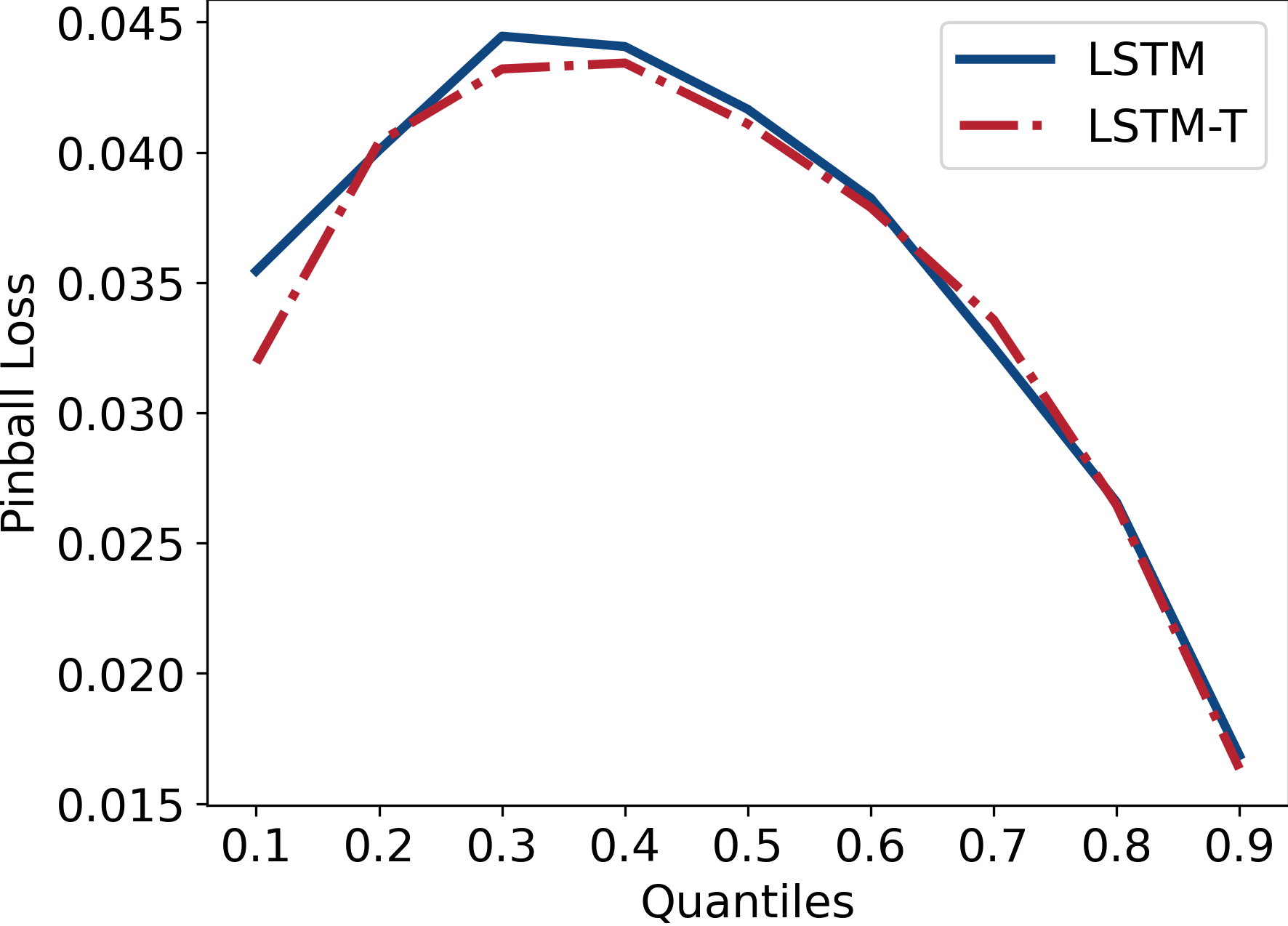}
  \caption{Pinball loss}
  \label{Pinball loss}
\end{subfigure}
\begin{subfigure}{0.3\textwidth}
  \includegraphics[width=\linewidth]{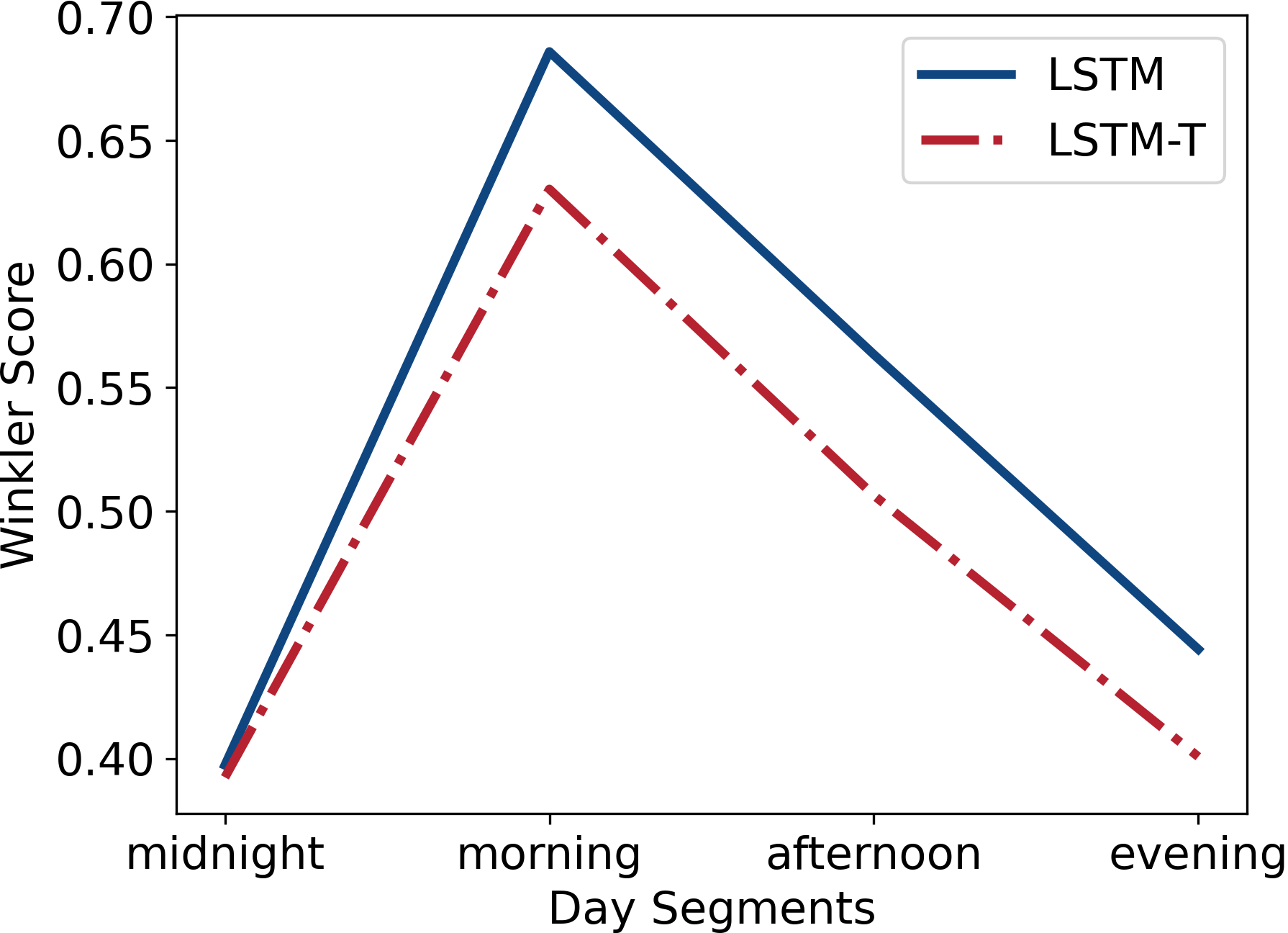}
  \caption{Winkler score}
  \label{Winkler score}
\end{subfigure}
\begin{subfigure}{0.3\textwidth}
  \includegraphics[width=\linewidth]{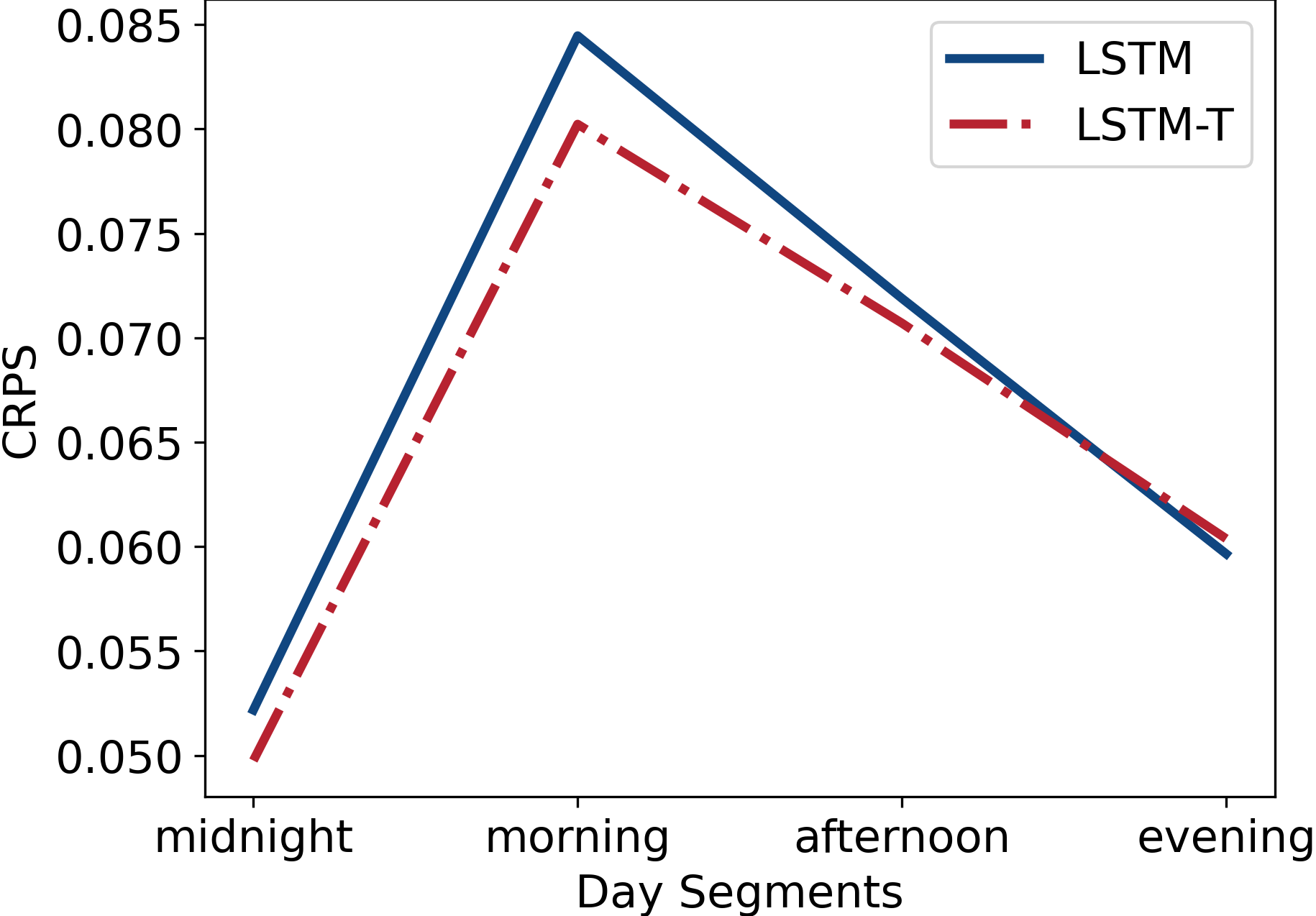}
  \caption{CRPS}
  \label{CRPS}
\end{subfigure}
\caption{Results of probabilistic forecasting}
\label{Results of probabilistic forecasting}
\end{figure*}

In the scenario of probabilistic forecasting, we evaluate the model performance by considering the quantile, the prediction interval (90\%), and the forecast distribution, to portray forecast uncertainty from the local to the global view.
The comparisons between the benchmark LSTM (blue solid lines) and the model with textual features (red dashed lines) are presented in Figure~\ref{Results of probabilistic forecasting}, which contains the errors of Pinball loss, Winkler score, and CRPS.

In Figure~\ref{Pinball loss}, the x-axis is the nine quantiles, and there is a slight improvement of LSTM with textual features on the lower quantiles.
The x-axes of Figure~\ref{Winkler score} and Figure~\ref{CRPS} are the day segments.
The prediction intervals in the whole day, except midnight, are narrowed by adding textual features.
Furthermore, the forecast distribution is also closer to the true distribution, and thus the LSTM with text information is more skillful and accurate in the morning.

\section{Conclusions}
\label{Conclusions}
As an extension of our previous work \cite{bai2023quantitative}, this paper explores the potential value of unstructured text information in electricity demand forecasting.
This work verifies again that the textual features benefit the LSTM forecasting paradigm.
In deterministic forecasting, the textual features assist in improving ~3\% from the LSTM benchmark, where the gaining is ~4\% in the ExtraTrees model.
Both the models with text are superior to the official benchmark from ENTSO-E.

Further research finds that the text features differ from those in \cite{bai2023quantitative} because the memory units in the LSTM model keep part of the history information when forecasting. However, the previous ExtraTrees-based regression model only takes the features from a day back.
In LSTM, the word frequency and topic distribution can no longer improve the forecasting, especially the Covid-19-related news that is useful in the ExtraTrees model but with limited assistance in LSTM.
Instead, public sentiment and more high-dimensional word representations in the news are of sustained influence, and the word representations involve information about transportation and geopolitical conflicts.
In addition, we use CNN autoencoder to obtain a more dense representation from word embeddings, improving the forecasting skills of the model.

To the best of our knowledge, few papers examine the impact of textual systems on probabilistic forecasting. 
We pioneer the field of electricity demand forecasting to explore whether textual information can reduce the uncertainty of forecasts.
The news reflects human social activity, and the experimental results reveal that effective textual features can narrow interval forecasts and bring the probabilistic forecast distribution closer to the true distribution. 
This phenomenon is more pronounced in the morning hours when human activity most influences electricity demand.

In conclusion, the study not only provides an example of combining the two fields of time series forecasting and natural language processing, but also can support future interdisciplinary collaborations between power systems and sociology. 
This study explores integrating knowledge from human activities in power systems to improve the effectiveness and social adaptability of smart grid applications.

\bibliographystyle{IEEEtran}
\bibliography{References.bib}

\begin{thebibliography}{10}
\providecommand{\url}[1]{#1}
\csname url@samestyle\endcsname
\providecommand{\newblock}{\relax}
\providecommand{\bibinfo}[2]{#2}
\providecommand{\BIBentrySTDinterwordspacing}{\spaceskip=0pt\relax}
\providecommand{\BIBentryALTinterwordstretchfactor}{4}
\providecommand{\BIBentryALTinterwordspacing}{\spaceskip=\fontdimen2\font plus
\BIBentryALTinterwordstretchfactor\fontdimen3\font minus \fontdimen4\font\relax}
\providecommand{\BIBforeignlanguage}[2]{{%
\expandafter\ifx\csname l@#1\endcsname\relax
\typeout{** WARNING: IEEEtran.bst: No hyphenation pattern has been}%
\typeout{** loaded for the language `#1'. Using the pattern for}%
\typeout{** the default language instead.}%
\else
\language=\csname l@#1\endcsname
\fi
#2}}
\providecommand{\BIBdecl}{\relax}
\BIBdecl

\bibitem{fallah2018computational}
S.~N. Fallah, R.~C. Deo, M.~Shojafar, M.~Conti, and S.~Shamshirband, ``Computational intelligence approaches for energy load forecasting in smart energy management grids: state of the art, future challenges, and research directions,'' \emph{Energies}, vol.~11, no.~3, p. 596, 2018.

\bibitem{metaxiotis2003artificial}
K.~Metaxiotis, A.~Kagiannas, D.~Askounis, and J.~Psarras, ``Artificial intelligence in short term electric load forecasting: a state-of-the-art survey for the researcher,'' \emph{Energy conversion and Management}, vol.~44, no.~9, pp. 1525--1534, 2003.

\bibitem{yang2019short}
A.~Yang, W.~Li, and X.~Yang, ``Short-term electricity load forecasting based on feature selection and least squares support vector machines,'' \emph{Knowledge-Based Systems}, vol. 163, pp. 159--173, 2019.

\bibitem{nystrup2020temporal}
P.~Nystrup, E.~Lindstr{\"o}m, P.~Pinson, and H.~Madsen, ``Temporal hierarchies with autocorrelation for load forecasting,'' \emph{European Journal of Operational Research}, vol. 280, no.~3, pp. 876--888, 2020.

\bibitem{thorey2018ensemble}
J.~Thorey, C.~Chaussin, and V.~Mallet, ``Ensemble forecast of photovoltaic power with online crps learning,'' \emph{International Journal of Forecasting}, vol.~34, no.~4, pp. 762--773, 2018.

\bibitem{browell2021probabilistic}
J.~Browell and M.~Fasiolo, ``Probabilistic forecasting of regional net-load with conditional extremes and gridded nwp,'' \emph{IEEE Transactions on Smart Grid}, vol.~12, no.~6, pp. 5011--5019, 2021.

\bibitem{sgarlato2022role}
R.~Sgarlato and F.~Ziel, ``The role of weather predictions in electricity price forecasting beyond the day-ahead horizon,'' \emph{IEEE Transactions on Power Systems}, 2022.

\bibitem{smith2010text}
N.~A. Smith, ``Text-driven forecasting,'' 2010.

\bibitem{li2019text}
X.~Li, W.~Shang, and S.~Wang, ``Text-based crude oil price forecasting: A deep learning approach,'' \emph{International Journal of Forecasting}, vol.~35, no.~4, pp. 1548--1560, 2019.

\bibitem{wu2021effective}
B.~Wu, L.~Wang, S.-X. Lv, and Y.-R. Zeng, ``Effective crude oil price forecasting using new text-based and big-data-driven model,'' \emph{Measurement}, vol. 168, p. 108468, 2021.

\bibitem{zhou2016can}
Z.~Zhou, J.~Zhao, and K.~Xu, ``Can online emotions predict the stock market in china?'' in \emph{Web Information Systems Engineering--WISE 2016: 17th International Conference, Shanghai, China, November 8-10, 2016, Proceedings, Part I 17}.\hskip 1em plus 0.5em minus 0.4em\relax Springer, 2016, pp. 328--342.

\bibitem{schaer2022predictive}
O.~Schaer, N.~Kourentzes, and R.~Fildes, ``Predictive competitive intelligence with prerelease online search traffic,'' \emph{Production and Operations Management}, vol.~31, no.~10, pp. 3823--3839, 2022.

\bibitem{wang2023forecasting}
D.~Wang, J.~Gan, J.~Mao, F.~Chen, and L.~Yu, ``Forecasting power demand in china with a cnn-lstm model including multimodal information,'' \emph{Energy}, vol. 263, p. 126012, 2023.

\bibitem{kozlowski2019geometry}
A.~C. Kozlowski, M.~Taddy, and J.~A. Evans, ``The geometry of culture: Analyzing the meanings of class through word embeddings,'' \emph{American Sociological Review}, vol.~84, no.~5, pp. 905--949, 2019.

\bibitem{bai2022crude}
Y.~Bai, X.~Li, H.~Yu, and S.~Jia, ``Crude oil price forecasting incorporating news text,'' \emph{International Journal of Forecasting}, vol.~38, no.~1, pp. 367--383, 2022.

\bibitem{sawhney2020multimodal}
R.~Sawhney, P.~Mathur, A.~Mangal, P.~Khanna, R.~R. Shah, and R.~Zimmermann, ``Multimodal multi-task financial risk forecasting,'' in \emph{Proceedings of the 28th ACM international conference on multimedia}, 2020, pp. 456--465.

\bibitem{rodrigues2019combining}
F.~Rodrigues, I.~Markou, and F.~C. Pereira, ``Combining time-series and textual data for taxi demand prediction in event areas: A deep learning approach,'' \emph{Information Fusion}, vol.~49, pp. 120--129, 2019.

\bibitem{liu2019combining}
J.~Liu, Z.~Lu, and W.~Du, ``Combining enterprise knowledge graph and news sentiment analysis for stock price prediction,'' 2019.

\bibitem{kreif2021estimating}
N.~Kreif, K.~DiazOrdaz, R.~Moreno-Serra, A.~Mirelman, T.~Hidayat, and M.~Suhrcke, ``Estimating heterogeneous policy impacts using causal machine learning: a case study of health insurance reform in indonesia,'' \emph{Health Services and Outcomes Research Methodology}, pp. 1--36, 2021.

\bibitem{joshi2010movie}
M.~Joshi, D.~Das, K.~Gimpel, and N.~A. Smith, ``Movie reviews and revenues: An experiment in text regression,'' in \emph{Human language technologies: The 2010 annual conference of the North American chapter of the Association for Computational Linguistics}, 2010, pp. 293--296.

\bibitem{obst2019textual}
D.~Obst, B.~Ghattas, S.~Claudel, J.~Cugliari, Y.~Goude, and G.~Oppenheim, ``Textual data for time series forecasting,'' \emph{arXiv preprint arXiv:1910.12618}, 2019.

\bibitem{obst2021textual}
D.~Obst, ``Textual data and transfer learning for time series forecasting,'' Ph.D. dissertation, Aix-Marseille, 2021.

\bibitem{bai2023quantitative}
Y.~Bai, S.~Camal, and A.~Michiorri, ``A quantitative exploration of natural language processing applications for electricity demand analysis,'' \emph{arXiv preprint arXiv:2301.07535}, 2023.

\bibitem{entso-e}
``{ENTSO-E. Transparency Platform},'' \url{https://transparency.entsoe.eu}.

\bibitem{BBC_news}
S.~Matthew, ``Bbc news front page archive,'' \url{https://dracos.co.uk/made/bbc-news-archive/archive.php}.

\bibitem{michelucci2022introduction}
U.~Michelucci, ``An introduction to autoencoders,'' \emph{arXiv preprint arXiv:2201.03898}, 2022.

\bibitem{hochreiter1997long}
S.~Hochreiter and J.~Schmidhuber, ``Long short-term memory,'' \emph{Neural computation}, vol.~9, no.~8, pp. 1735--1780, 1997.

\bibitem{Rob2021Evaluating}
H.~Rob, J and A.~George, ``Evaluating distributional forecast accuracy,'' in \emph{Forecasting: Principles and Practice}.\hskip 1em plus 0.5em minus 0.4em\relax OTexts: Melbourne, Australia. OTexts.com/fpp3, 2021.

\end{thebibliography}

\end{document}